%% file: main.tex
\documentclass{article}
\usepackage{spconf,amsmath,graphicx}
\usepackage{amssymb}
\usepackage{algorithm,algpseudocode}   
\usepackage{multirow}
\usepackage{xcolor}        
\usepackage{mathtools}
\usepackage[mathscr]{euscript}

\input{definitions}

\title{Compressive Sensing with Tensorized Autoencoder}

\name{Rakib Hyder and M. Salman Asif\sthanks{This work was supported in part by ONR N00014-19-1-2264 and AFOSR FA9550-21-1-0330 grants.}}
\address{University of California Riverside}

\begin{document}

\maketitle

\begin{abstract}

Deep networks can be trained to map images into a low-dimensional latent space. In many cases, different images in a collection are articulated versions of one another; for example, same object with different lighting, background, or pose. Furthermore, in many cases, parts of images can be corrupted by noise or missing entries. In this paper, our goal is to recover images without access to the ground-truth (clean) images using the articulations as structural prior of the data. Such recovery problems fall under the domain of compressive sensing. We propose to learn autoencoder with tensor ring factorization on the the embedding space to impose structural constraints on the data. In particular, we use a tensor ring structure in the bottleneck layer of the autoencoder that utilizes the soft labels of the structured dataset. We empirically demonstrate the effectiveness of the proposed approach for inpainting and denoising applications. The resulting method achieves better reconstruction quality compared to other generative prior-based self-supervised recovery approaches for compressive sensing.
\end{abstract}

\section{Introduction}
\label{sec:intro}
Low-rank tensor factorization is a powerful tool to represent multi-dimensional and multi-modal data using a small number of low-dimensional factors (cores) \cite{kolda2009tensor}. 
Tensor factorization has also been recently used for compressing data and neural network parameters  \cite{bacciu2020tensor,ji2019survey}. 
Deep autoencoders and generative models, such as generative adversarial networks (GAN) \cite{goodfellow2014generative}, variational autoencoders (VAE) \cite{kingma2013auto}, and generative latent optimization (GLO) \cite{Bojanowski2018OptimizingTL}, also provide an excellent mechanism to learn low-dimensional representation of data.

In this paper, we combine tensor factorization with an autoencoder to recover a collection of articulated images from their corrupted or compressive measurements. Such articulated images often arise in surveillance and multi-view sensing applications  \cite{vlasic2006face,deng2017factorized,zhang2019robust}. Images with different imperfections (or their indirect measurements) can be modeled as
\begin{equation}
    y_i=\mathbf{A}_ix_i+\eta_i,
    \label{eq:measModel}
\end{equation}
where $x_i$ denotes the $i^{th}$ image, $y_i$ denotes the observed measurements, $\mathbf{A}_i$ denotes the corresponding measurement matrix (corruption model), and $\eta_i$ denotes the corresponding measurement noise. 

Our main goal is to recover the structured images \{$x_i$\} from the available measurements without access to the ground-truth (clean) images. To achieve this goal, we use the known image articulations as structural prior of the data.  
In particular, we learn an encoder that maps every measurement to a latent space. We represent the latent codes as a low-rank tensor, where different factors represent different image articulations. We learn a decoder that maps the low-rank tensor to the images. 
We present several experiments to demonstrate that our proposed method outperforms other self-supervised methods that use low-rank tensors or generative priors.

\begin{figure*}[ht]
    \centering
    \includegraphics[width=0.8\linewidth]{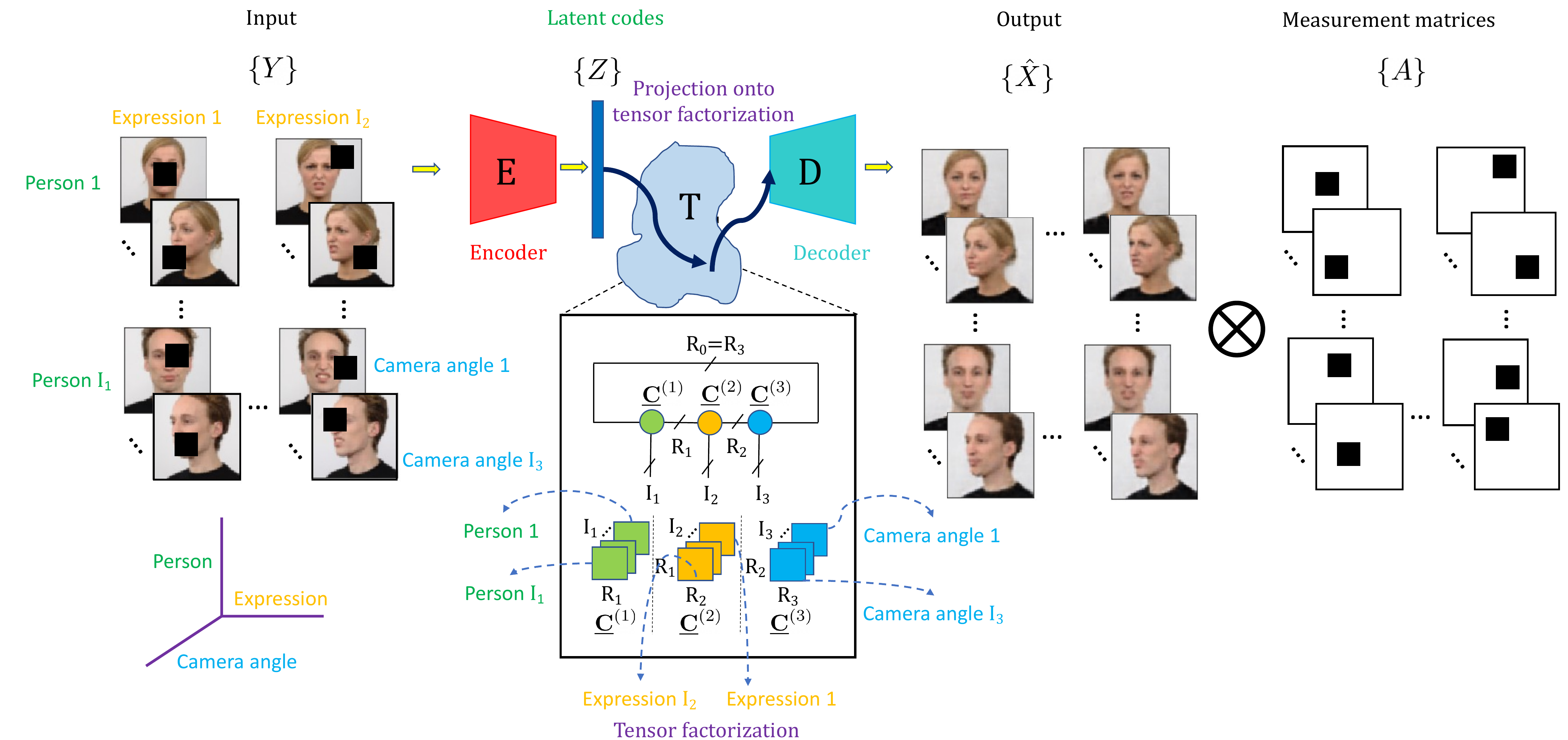}
    \caption{General overview of our proposed tensor ring factorized autoencoder. We map a set of measurements $\{Y\}$ to latent codes $\{Z\}$ using an encoder $E$. We then perform tensor factorization on the latent space codes using tensor factorization (shown as $\mathbf{T}$ block). Finally, we pass the factorized representation though the decoder $D$ to generate target images $\hat X$. } 
    \label{fig:intro}
\end{figure*}

\subsection{Related Work}
\noindent {\bf Tensor factorization} has been a  popular method for multidimensional data and complex network compression \cite{wang2018wide,tjandra2018tensor}. Different tensor factors can be used to represents different attributes of the data and potentially generate novel/missing data  \cite{chen2019bayesian,krishnaswamy2019tensor}. 

We are using the strength of tensor factorization to compress the latent space of the deterministic autoencoder using explicit low-rank constraints in order to use it as a generative prior. We are also using it to factorize the representation of structured set of images in the latent space to achieve better recovery  performance.
   
\noindent {\bf Compressive sensing} refers to a broad class of problems in which we aim to recover a signal from a small number of measurements \cite{candes2011compressed, donoho2006compressed, candes2005decoding}. The canonical problem in \eqref{eq:measModel} can be underdetermined, and we need to use some prior knowledge about the signal structure. Classical signal priors exploit sparse and low-rank structures in images and videos for their reconstruction \cite{baraniuk2007compressive, yang2013adaptive,zhao2017video,asif2013low}. 
However, the natural images exhibits far richer nonlinear structures than sparsity alone. 
We focus on a newly emerging family of data-driven representation methods based on generative models that are either learned from training data or measurements \cite{bora2017compressed,heckel2018deep,Ulyanov2017DeepIP}. 
A number of methods optimize generator network weights while keeping the latent code fixed at a random value  \cite{heckel2018deep,van2018compressed}. Both DIP \cite{Ulyanov2017DeepIP} and deep decoder \cite{heckel2018deep} update the network parameters to generate a given image; therefore, the generator can reconstruct wide range of images.

\section{Technical Details}

Given corrupted or compressed measurements $\{y_i\}$ and  measurement matrices $\{\mathbf{A_i}\}$, our aim is to estimate the true image collection $\{x_i\}$. We learn a deterministic autoencoder by solving the following constrained optimization problem:
\begin{equation}
    \underset{\theta,\gamma}{min}\; \sum_{i=1}^{N} \|y_i-\mathbf{A_i}D(E(y_i;\theta);\gamma)\| \; \text{s.t. }\; E(y_i;\theta) \in \mathbb{T}, \label{eq:min}
\end{equation}
\noindent where $E(\cdot;\theta)$ denotes an encoder, $D(\cdot;\gamma)$ denotes a decoder, $\mathbb{T}$ represents the tensor-factorized latent space, and $N$ denotes the total number of samples in the target set. For any measurement $y_i$, we denote the latent code representation as $z_i=E(y_i;\theta)\in \mathbb{R}^d$.

Tensor factorization can represent multi-dimensional and multi-modal data using a small number of low-dimensional factors. Instead of applying the tensor factorization directly on the image/signal space, we factorize the low-dimensional latent space. We seek two main goals with such factorization: 1) limit the degrees of freedom for the latent space and 2) utilize the structural similarity of the dataset for the self supervised compressive sensing. 

We denote the $N$ latent codes of our entire target set as $\{z_1,  \ldots, z_N\}$. For a structured dataset, these latent codes can be factored into $K$ different attributes, each of which has $I_k$ variants for $k \in \{1,2, \ldots, K\}$; therefore, we can write $N=I_1\times I_2 \times \ldots \times I_K$. We  denote the latent code tensor with all the $z_i$ as $\mathscr{Z}$, which is an $I_1\times I_2 \times \ldots \times I_K\times d$ tensor, and $\mathscr{Z}(i_1,\ldots, i_K)$ denotes one of the $d$-dimensional latent codes and $\mathscr{Z}(i_1,\ldots, i_K, i_{K+1})$ denotes one of the entries of a $d$-dimensional latent code. 

We use tensor ring factorization in this paper. Brief description for tensor ring factorization is presented below. For detailed discussion, we refer the readers to \cite{yokota2016tensor}. 

A \textit{tensor ring} (TR) decomposition can represent a latent tensor $\mathscr{Z}$ using $K+1$ different 3-order tensor cores : $\umC^{(1)}, \ldots, \umC^{(K+1)}$ representing $K$ different attributes and $d$ dimensional code, where $\umC^{(k)}  \in \bR^{R_{k-1}\times I_k \times R_k}$ and $(R_1,\ldots, R_{K+1})$ denotes the multilinear rank with $R_0=R_{K+1}$. All the entries in $\mathscr{Z}$ can be represented as 
\begin{equation}
        \mathscr{Z}(i_1,\ldots, i_{K+1}) = \sum_{R_1, \ldots, R_{K+1}} \prod_{j=1}^{K+1}\umC^{(j)}(R_{j-1},i_j,R_j), \label{eq:TR}
\end{equation}

where $\umC^{(k)}(:,i_k,:)$ denotes $i_k$th slice of $\umC^{(k)}$ that is an $R_{k-1}\times R_k$ matrix and the trace operation sums up all the diagonal entries. 

The total number of elements in $\mathscr{Z}$ is $d\prod_{k = 1}^K I_k$. The total number of parameters in TR factorization reduces to $\sum_{k=1}^K I_k R_{k-1}R_k+d R_{K}R_{0}$ with $R_0 = R_{K+1}$. If we set all the $R_k = R$, then the total number of parameters in TR factorization becomes $R^2(\sum_k I_k+d)$, which is significantly less than $d \prod_k I_k$ in $\mathscr{Z}$.

We modify the optimization problem in \eqref{eq:min} to use the following loss function:
\begin{equation}
\begin{split}
  \text{Loss} =&\sum_{i=1}^{N} \|E(y_i;\theta)-\mathscr{Z}(i_1,\ldots i_K)\|^2\\
  &+\lambda_1\|\mathbf{A_i}D(E(y_i;\theta);\gamma)-y_i\|^2 \\
  &+\lambda_2\|\mathbf{A_i}D(\mathscr{Z}(i_1,\ldots i_K);\gamma)-y_i\|^2. 
\end{split}
    \label{eq:miss_loss}
\end{equation}
The three terms in the loss function in \eqref{eq:miss_loss} are targeted to minimize the mismatch between encoder output and factorization, encoder-decoder measurement loss and factorization-decoder measurement loss respectively. $\lambda_1$ and $\lambda_2$ are weights for different loss terms. The first term of the total loss in \eqref{eq:miss_loss} measures  the mismatch between encoder output and factorization. 
The second term of the total loss in \eqref{eq:miss_loss} evaluates how well the encoder output performs in terms of reconstruction.
The third term in \eqref{eq:miss_loss} measures how well the decoder performs when given the latent codes formed by the tensor factors. Note that if the first term were perfectly zero, the third term would not be necessary. Even though the output from the encoder and the output from the tensor factors are very close, they may provide very different realizations when passed through decoder depending on the direction of mismatch. We kept an extra term to make sure that the latent representation from the tensor factorization also generates as good images as encoder output does. We present a pseudocode for the recovery algorithm in Algorithm~\ref{algo:miss}.

 \begin{algorithm}[tb]
  \caption{Learning Tensor Factorization using Autoencoder from Corrupted Data}
  \label{algo:miss}
 \begin{algorithmic}
  \State {\bfseries Input:} Measuremnets $\{y_i\}$,measurement model $\bf{A}_i$,  and attribute label of the data.

  \State Initialize encoder and decoder weights and tensor cores randomly
 	\For{$m = 1, 2, \ldots, M$}    \Comment{$M$ steps or until convergence}
        \State Calculate the loss function in \eqref{eq:miss_loss}.
     	\State Calculate gradients of loss w.r.t. training parameters in $\mathbb{T}$, $E(.;\theta)$ and $D(.;\gamma)$  via backpropagation.
     	\State Update parameters using gradient descent.
 	\EndFor
  \State {\bfseries Intermediate Output:} Optimized $\mathbb{T}$, $E(\cdot;\theta)$ and $D(\cdot;\gamma)$.
  \State Use optimized $\mathbb{T}$ and $D(\cdot;\gamma)$ to estimate the unknown signal as $\hat x_i=D(z_i;\gamma)$ where $z_i\in \mathbb{T}$.
  \State {\bfseries Output:} $\{x_i\}$

 \end{algorithmic}
 \end{algorithm}

\begin{figure*}[!ht]
    \centering
        \includegraphics[width=0.9\textwidth]{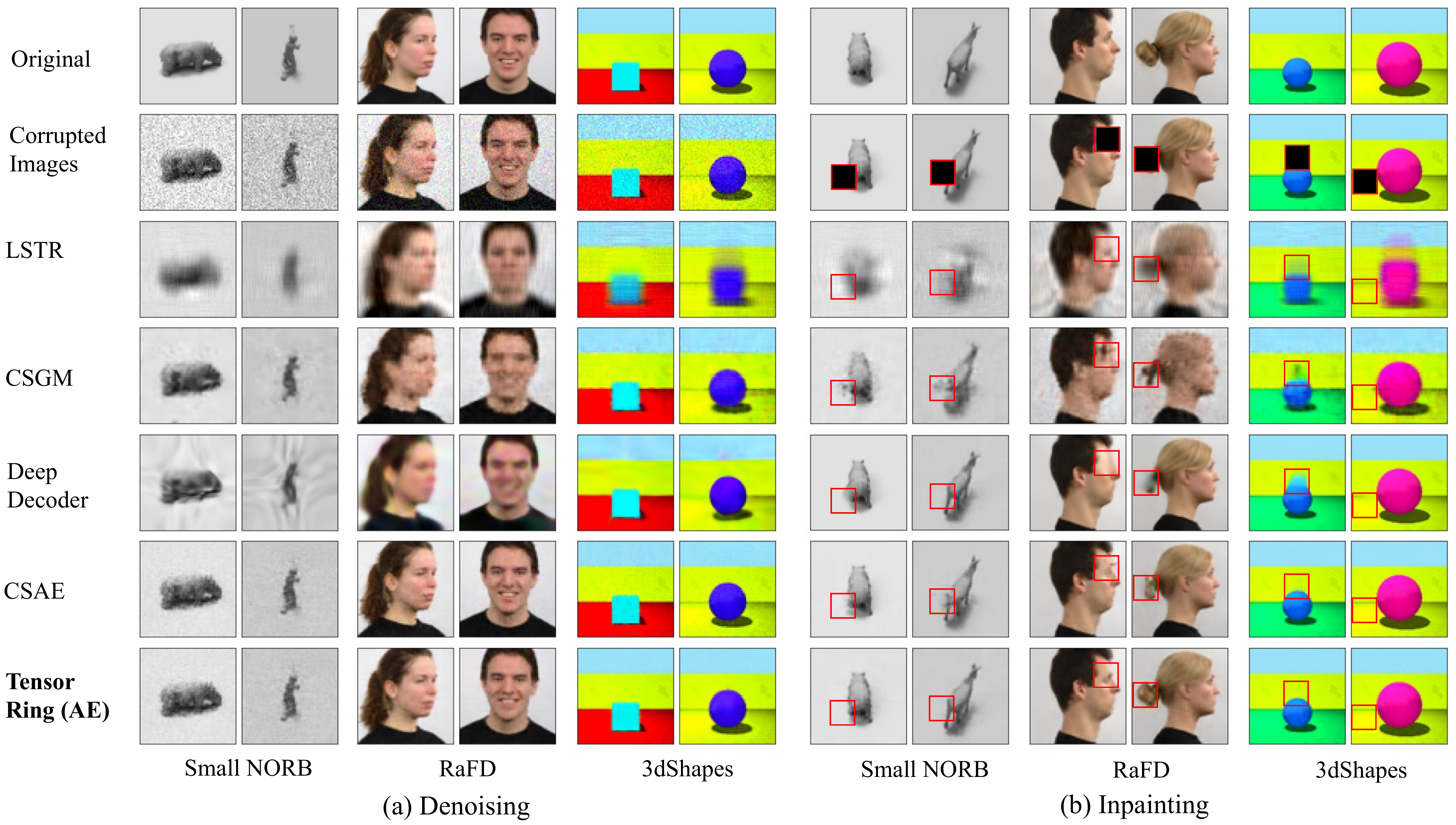}

    \caption{Reconstruction results for (a) denoising and (b) inpainting on Small NORB, RaFD and 3dShapes datasets.}
    \label{fig:rec}
    \vspace{-3ex}
\end{figure*}

\section{Experiments and Results}

\noindent \textbf{Dataset:} We used Small NORB \cite{lecun2004learning}, RaFD \cite{langner2010presentation} and 3dShapes \cite{3dshapes18} datasets in our experiments. In these datasets, images of different attribute variation is available. We select 25 toys with 3 lighting conditions, 3 elevations and 9 azimuth angles (2025 images) from Small NORB dataset. We select 15 persons with 5 camera angles, 8 expressions and 3 eye gazing (1800 images) from RaFD. We selected 4 object shapes, 5 floor colors and 5 floor colors at 8 object scales (800 images) from 3dShapes dataset.

 \noindent \textbf{Setup:} In our experiments, we used a fully convolutional autoencoder that maps each image to a latent code $z\in \mathbb{R}^{16\times4\times4}$, which sets $d=256$ in our experiments. Our encoder consists of four convolutional layers (32,64,128,16 filters) and decoder consists of four transpose convolutional layers (256, 128,64,3 (rgb) /1 (grayscale) filters) each with $3\times 3$ filters with stride=2 followed by ReLU activation except for the last layer (that uses Sigmoid instead of ReLU). We use low-rank tensor ring factorization in the latent space. We use the same rank for all the cores of tensor ring. We empirically selected the lowest ranks for each dataset that provide good performance. We reported results with rank=25 for Small NORB, 30 for RaFD and 15 for 3dShapes.  We have initialized the cores and bases for different tensor factorization using samples drawn from $N(0,0.1)$ distribution. We kept fixed seed for every setup in order to achieve fair comparison. As we consider a batch of images correspond to a slice of a tensor core, we could perform minibatch optimization which reduced memory requirement during training. We have used Adam \cite{kingma2014adam} optimization for network parameters optimization and Stochastic Gradient Descent (SGD) for optimizing tensor factorization parameters. The learning rate for Adam was selected to be $0.001$ and SGD to be $1$ (or $0.1$). We let the optimization run for enough iterations to converge. 
 For autoencoder setup, we set weight terms, $\lambda_1$ and $\lambda_2$ to be $1$.

 \begin{table}[!ht]
\centering
\caption{Reconstruction quality (PSNR in dB) for image denoising and inpainting with different comparing approaches.}
\label{table:rec}
\small
\begin{tabular}{llllll}
\hline
\multicolumn{1}{|l|}{}          & \multicolumn{1}{l|}{LSTR}  & \multicolumn{1}{l|}{\begin{tabular}[c]{@{}l@{}}Deep\\ Decoder\end{tabular}} & \multicolumn{1}{l|}{CSGM}  &\multicolumn{1}{l|}{CSAE}& \multicolumn{1}{l|}{\begin{tabular}[c]{@{}l@{}}Tensor \\ Ring AE\end{tabular}} \\ \hline
\hline
\multicolumn{5}{c}{Denoising}                                                                                                                                                                                                                                                                                      \\ \hline \hline
\multicolumn{1}{|l|}{Small NORB} & \multicolumn{1}{l|}{23.11}                                                          & \multicolumn{1}{l|}{27.40}                                                   & \multicolumn{1}{l|}{28.40}  &\multicolumn{1}{l|}{27.56}  &\multicolumn{1}{l|}{\textbf{31.71}}                                            \\ \hline
\multicolumn{1}{|l|}{RaFD}       & \multicolumn{1}{l|}{20.30}                                                           & \multicolumn{1}{l|}{25.14}                                                  & \multicolumn{1}{l|}{26.67} &\multicolumn{1}{l|}{29.66}  &\multicolumn{1}{l|}{\textbf{32.10}}                                            \\ \hline
\multicolumn{1}{|l|}{3dShapes}   & \multicolumn{1}{l|}{20.13}                                                          & \multicolumn{1}{l|}{28.06}                                                  & \multicolumn{1}{l|}{28.63} &\multicolumn{1}{l|}{33.52}  &\multicolumn{1}{l|}{\textbf{35.97}}                                             \\ \hline \hline
\multicolumn{5}{c}{Inpainting}                                                                                                                                                                                                                                                                                     \\ \hline \hline
\multicolumn{1}{|l|}{Small NORB} & \multicolumn{1}{l|}{24.70}                                                               & \multicolumn{1}{l|}{35.10}                                                       & \multicolumn{1}{l|}{28.00}   &\multicolumn{1}{l|}{33.49}    & \multicolumn{1}{l|}{\textbf{35.29}}                                                          \\ \hline
\multicolumn{1}{|l|}{RaFD}       & \multicolumn{1}{l|}{21.31}                                                               & \multicolumn{1}{l|}{31.87}                                                       & \multicolumn{1}{l|}{24.83}      &\multicolumn{1}{l|}{31.91}  &\multicolumn{1}{l|}{\textbf{33.55} }                                                         \\ \hline
\multicolumn{1}{|l|}{3dShapes}   & \multicolumn{1}{l|}{21.92}                                                               & \multicolumn{1}{l|}{35.22}                                                       & \multicolumn{1}{l|}{27.26}      &\multicolumn{1}{l|}{36.8}  &\multicolumn{1}{l|}{\textbf{39.43}}                                                         \\ \hline
\end{tabular}

\end{table}

\noindent \textbf{Comparison:} We solve the inverse/compressive sensing problems (e.g. denoising and inpainting) without ground truth images. We show comparison with 4 different baselines which also perform the same task.

\noindent \textbf{CSAE:}  We use the encoder of a deterministic autoencoder to learn the latent space from the corrupted measurements and pass the learned latent codes through decoder and measurement matrices to match with the observed measurements. One can refer to the second term of Eqn~\eqref{eq:miss_loss} as the objective this approach minimizes. Eventually we learn the original image given the corrupted measurements without having the ground truth. We term it \textbf{CSAE} (Compressive Sensing with Auto Encoder).

\noindent \textbf{CSGM:} We tried to solve the compressive sensing problems given that we have a trained generative model with the learned distribution of the target data. It is similar to the work of Bora et. al. \cite{bora2017compressed}. We term it \textbf{CSGM} following \cite{bora2017compressed}.

\noindent \textbf{LSTR:} We utilize the attribute information of the structured dataset and use it as a prior to minimize the least square measurement loss with SGD. We term it \textbf{LSTR} (Least Square minimization with Tensor Ring).

\noindent \textbf{Deep Decoder:} Finally we use one of the self supervised generative prior based approaches to solve the compressive sensing problems. We use \textbf{Deep Decoder} \cite{heckel2018deep} for comparison.

We empirically demonstrate that our proposed Tensor Ring factorized Autoencoder outperforms all the four baselines in terms of reconstruction quality since we are using the advantages of both the structural information and generative priors.
\\

 \noindent \textbf{Denoising:} In this experiment, we added Gaussian noise of 20dB to all the images. We report the average reconstruction quality (dB PSNR) for different comparing techniques in Table~\ref{table:rec}. We also demonstrate some reconstructed images in Figure~\ref{fig:rec}. We can observe that utilizing the structure in latent space helps us outperform the other approaches. Deep deocoder uses a single network per image recovery. So it cannot use information from the other measurements of the structure. Although CSGM uses all the training data to train the generator, it does not explicitly use the structural information. We also observe that LSTR does not provide good reconstruction performance even though it is also using the structural information because images usually do not have the tensor structure in their representation. CSAE approach performs well as it learns the optimal embedding space for solving the inverse problem using an encoder. However, it falls behind our proposed approach since it does not use any structural information. By learning an embedding space to apply tensor structure, we are utilizing the structural information to our advantage.
 
 \noindent \textbf{Image Inpainting:} It is often observed in real scenario that some of the images of the structured image set are corrupted instead of being  completely unavailable. We perform a set of experiments on different datasets where we missed a $16\times16$ block from all the images at random locations. We feed the structured image set to the AE based tensor factorized scheme. Our tensor factorized autoencoder utilize the strength of the structured organization of the dataset to better reconstruct the images with missing blocks. We report the reconstruction results in Table~\ref{table:rec}. We also demonstrate some reconstructions in Figure~\ref{fig:rec}. We can observe that we outperform the other approaches especially in recovering the original details of missing blocks as shown in Figure~\ref{fig:rec}. Although Deep Decoder and CSAE perform very close to our approach, they fail to recover reliable details in the missing blocks. Since deep decoder uses a separate network for every image, its memory and parameter requirements are significantly higher than our method.

\section{Conclusion}
We proposed tensorized autoencoder as a prior for solving compressive sensing problems. For structured datasets, we utilize the structural similarities in the images by applying tensor ring factorization in the latent space learned by an encoder. We demonstrated that applying structural constraint such as tensor ring performs better on the learned latent space. We also observed that by utilizing the structural similarity of the dataset, tensorized autoencoder can outperform other self supervised generative prior and deep image prior based approaches for different compressive sensing applications.

\ninept 
\bibliographystyle{IEEEbib}
\bibliography{main}

\end{document}

%% file: definitions.tex
\newcommand{\umC}{\ensuremath{\underline{\mathbf{C}}}}

\newcommand{\bR}{\ensuremath{\mathbb{R}}}